\documentclass{article}

\usepackage{microtype}
\usepackage{graphicx}
\usepackage{subfigure}
\usepackage{booktabs} 

\usepackage{hyperref}

\usepackage[accepted]{icml2025}

\usepackage{amsmath}
\usepackage{amssymb}
\usepackage{mathtools}
\usepackage{amsthm}

\usepackage[capitalize,noabbrev]{cleveref}

\theoremstyle{plain}

\theoremstyle{definition}

\theoremstyle{remark}

\usepackage[textsize=tiny]{todonotes}

\icmltitlerunning{MultiNet: An Open-Source Software Toolkit \& Benchmark Suite for the Evaluation and Adaptation of Multimodal Action Models}

\begin{document}

\twocolumn[
\icmltitle{An Open-Source Software Toolkit \& Benchmark Suite for the Evaluation and Adaptation of Multimodal Action Models}



\icmlsetsymbol{equal}{*}

\begin{icmlauthorlist}
\icmlauthor{Pranav Guruprasad}{equal,yyy,xxx}
\icmlauthor{Yangyue Wang}{equal,yyy,xxx}
\icmlauthor{Sudipta Chowdhury}{xxx}
\icmlauthor{Jaewoo Song}{xxx}
\icmlauthor{Harshvardhan Sikka}{yyy,xxx,zzz}
\end{icmlauthorlist}

\icmlaffiliation{yyy}{Metarch}
\icmlaffiliation{xxx}{Manifold Research}
\icmlaffiliation{zzz}{Georgia Institute of Technology}

\icmlcorrespondingauthor{Pranav Guruprasad}{pranav@metarch.ai}

\icmlkeywords{Machine Learning, ICML}

\vskip 0.3in
]



\printAffiliationsAndNotice{}  

\begin{abstract}
Recent innovations in multimodal action models represent a promising direction for developing general-purpose agentic systems, combining visual understanding, language comprehension, and action generation. We introduce MultiNet - a novel, fully open-source benchmark and surrounding software ecosystem designed to rigorously evaluate and adapt models across vision, language, and action domains. We establish standardized evaluation protocols for assessing vision-language models (VLMs) and vision-language-action models (VLAs), and provide open source software to download relevant data, models, and evaluations. Additionally, we provide a composite dataset with over 1.3 trillion tokens of image captioning, visual question answering, commonsense reasoning, robotic control, digital game-play, simulated locomotion/manipulation, and many more tasks. The MultiNet benchmark, framework, toolkit, and evaluation harness have been used in downstream research on the limitations of VLA generalization.
\end{abstract}
\section{Introduction}
Recent advances in machine learning have demonstrated the potential of large-scale models to exhibit broad generalization capabilities across diverse tasks. Vision-Language-Action (VLA) models in particular have showcased impressive abilities in grounding real-world actions with visual perception and natural language understanding \cite{openvla2024} \cite{pi02024} \cite{pi0fast2025}. These models can interpret complex visual scenes, comprehend natural language commands, and generate appropriate control sequences. Nevertheless, contemporary VLA models are predominantly engineered for, and evaluated on, narrow domains such as robotic manipulation \cite{guruprasad2024benchmarkingvisionlanguage}. Their efficacy on multifaceted vision-language understanding tasks or their capacity for broader generalization remains largely unproven, a limitation potentially stemming from training regimens that prioritize specific capabilities over holistic, generalist behavior.

Developing functional agents capable of generalizing to a broad set of action tasks necessitates training on vast, diverse datasets encompassing multiple modalities (vision, language, control) and a wide array of task types. Such agents must not only excel within individual modalities but also seamlessly integrate information and execute actions across them, mirroring the multifaceted demands of real-world scenarios \cite{guruprasad2025benchmarkingvisionlanguage}. Presently, the community lacks a large-scale, open-source benchmark specifically architect ed for the rigorous training and comprehensive evaluation of these ambitious generalist models. Most benchmarks focus on narrow, very specific domains and tasks, and are typically closed-source \cite{white2024livebenchchallengingcontaminationfreellm} \cite{hendrycks2021measuringmassivemultitasklanguage} \cite{liang2021multibenchmultiscalebenchmarksmultimodal} \cite{gulcehre2021rlunpluggedsuitebenchmarks} \cite{fu2021d4rldatasetsdeepdatadriven}. This critical gap motivates our work.

In this paper, we introduce the MultiNet ecosystem - a comprehensive set of benchmarking software designed to catalyze the development and evaluation of generalist action models. Our contributions are multifaceted:

\begin{itemize}
    \item \textbf{A Large-Scale Generalist Dataset}: Release of an extensive, Open-Source dataset that amalgamates diverse data sources (vision, language, and action) suitable for training and evaluating generalist models as seen in \ref{dataset_spec}. 
    \item \textbf{An Open-Source Data Curation SDK}: Open-source access to a software toolkit to facilitate easy access to the consolidated dataset. This toolkit also standardizes control data (Reinforcement Learning and Robotics data) from a myriad of sources into a common, accessible format.
    \item \textbf{A Systematic Evaluation Harness}: A standardized, well-reasoned evaluation methodology including test splits and carefully designed metrics as seen in \ref{metrics}. This framework is specifically engineered to provide easy access to the community to assess the generalization capabilities of state-of-the-art Vision-Language Models (VLMs) and Vision-Language-Action (VLA) models across a spectrum of familiar and novel domains, including real-world robotics tasks, and procedurally generated game environments.
    \item \textbf{Open-source adaptations for SoTA VLA models and VLMs}:
      Open-sourced adaptations of SoTA VLMs and VLAs, enabling them to operate effectively on the data formats and diverse domains in MultiNet, even those unseen during their original training, to further accelerate progress toward building generalist AI systems.
    \item \textbf{In-depth Experiments and Analysis}: Utilization of the MultiNet benchmark, framework,evaluation harness, and model adaptations to obtain evaluations and analysis of the performance of leading VLMs, VLAs, and emerging generalist models.
\end{itemize}

Through MultiNet, we aim to provide the community with the essential resources—datasets, tools, and standardized evaluation protocols—to systematically compare different approaches, gain deeper insights into the challenges and opportunities in building generalist AI, and ultimately accelerate the development of truly general-purpose intelligent systems.

\section{A Large-scale Open-Source Generalist Dataset}
\begin{figure}
    \centering
    \includegraphics[width=1\linewidth]{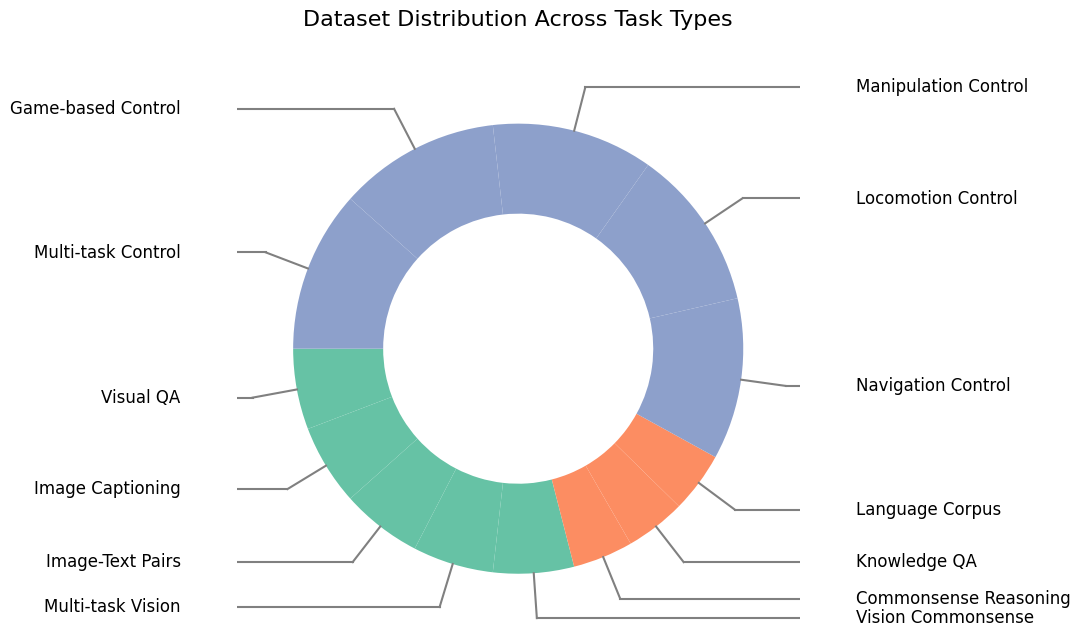}
    \caption{Control represents the largest portion (58\%) due to the extensive OpenX-Embodiment collection, followed by Vision-Language (29\%) and Language (13\%) datasets}
    \label{fig:datasets_pi_chart}
\end{figure}
Through this work, we provide a comprehensive dataset collection to facilitate the training and evaluation of generalist AI models, and fully open-source easy access to it. Our collection aggregates a diverse range of existing datasets, encompassing various domains such as vision-language understanding, language processing, reinforcement learning, and robotics. This collection provides a rich resource by bringing together data across multiple domains, modalities and tasks, aiming to support the development of more capable and versatile AI systems.

The collection includes a variety of datasets focused on different aspects of vision-language understanding: OBELICS \cite{laurençon2023obelicsopenwebscalefiltered}, an open web-scale dataset of 141 million interleaved image-text web pages, 353 million images, and 115 billion text tokens; DataComp-1B \cite{gadre2023datacompsearchgenerationmultimodal}, a curated set of 1.4 billion image-text pairs; COYO-700M \cite{coyo}, containing 747 million image-alt-text pairs with minimal filtering; MS-COCO Captions \cite{lin2015microsoftcococommonobjects}, featuring 330,000 images each with 5 captions; Conceptual Captions \cite{sharma-etal-2018-conceptual}, consisting of 3.3 million web-harvested images with filtered descriptions; A-OKVQA, \cite{schwenk2022aokvqabenchmarkvisualquestion} with 24,903 question/answer/rationale triplets requiring broad commonsense knowledge; VQA-V2 \cite{goyal2017makingvvqamatter}, offering 265,000 images with open-ended questions that demand understanding of vision, language, and commonsense; Flickr30k \cite{plummer2016flickr30kentitiescollectingregiontophrase}, with 31,000 images paired with five human-annotated sentences; TextVQA \cite{singh2019towards}, containing 45,336 questions on 28,408 images to assess reading and reasoning about text within images; VizWiz \cite{gurari2018vizwizgrandchallengeanswering}, a collection of images taken by blind individuals with associated questions; WinoGAViL \cite{bitton2022winogavilgamifiedassociationbenchmark}, which tests vision-and-language commonsense reasoning; ImageNet-R \cite{hendrycks2021facesrobustnesscriticalanalysis}, featuring artistic renditions of 200 ImageNet classes; ObjectNet \cite{NEURIPS2019_97af07a1}, a real-world test set for object recognition with random backgrounds, rotations, and viewpoints;

Further enriching the collection are datasets focused on reinforcement learning, and robotics tasks: DM Lab \cite{beattie2016deepmindlab}, providing frames from the DeepMind Lab environment annotated with agent-object distances; ALE Atari \cite{Bellemare_2013}, with 57 Atari 2600 game environments and 500,000 interactions per game; BabyAI \cite{chevalierboisvert2019babyaiplatformstudysample}, a platform with 19 difficulty levels and 100,000 episodes for grounded language learning ; MuJoCo \cite{mujoco}, a benchmark suite of 11 continuous control tasks with 10,000 episodes per environment; DM Control Suite \cite{tassa2018deepmindcontrolsuite}, offering standardized continuous control environments; V-D4RL \cite{lu2023challengesopportunitiesofflinereinforcement}, a benchmark for continuous control from visual observations of DM Control Suite tasks ; Meta-World \cite{yu2021metaworldbenchmarkevaluationmultitask}, providing the MT50 benchmark with 50 diverse robot manipulation tasks and 10,000 episodes per environment; Procgen \cite{cobbe2020leveragingproceduralgenerationbenchmark}, OpenAI’s suite of 16 procedurally generated game-like environments; OpenX-Embodiment \cite{open_x_embodiment_rt_x_2023}, the largest open-source real robot dataset with over 1 million trajectories from 22 embodiments; and LocoMuJoCo \cite{alhafez2023locomujococomprehensiveimitationlearning}, an imitation learning benchmark for locomotion using real noisy motion capture data.

The consolidated set also incorporates datasets that can help train and evaluate systems on advanced language capabilities: Fineweb-edu \cite{penedo2024finewebdatasetsdecantingweb}, providing 1.3 trillion tokens of educational content filtered from the FineWeb dataset; HellaSwag \cite{zellers2019hellaswagmachinereallyfinish}, comprising 70,000 multiple-choice questions for commonsense natural language inference ; ARC (AI2 Reasoning Challenge)\cite{clark2018thinksolvedquestionanswering}, providing science exam questions from grades 3-9 ; CommonsenseQA \cite{talmor2019commonsenseqaquestionansweringchallenge}, containing 12,247 multiple-choice questions requiring general world knowledge ; and MMLU \cite{hendrycks2021measuringmassivemultitasklanguage}, a benchmark evaluating language models across 57 subjects with approximately 16,000 multiple-choice questions; 

\section{Open-Source Dataset SDK}
Our open-source codebase \footnote{https://github.com/ManifoldRG/MultiNet} allows for the seamless download of any or all datasets in our collection, along with a toolkit for standardizing robotics and reinforcement learning data. Addressing the common issues of outdated formats, poor maintenance, and accessibility challenges found in many existing control datasets, our toolkit provides stable access methods for diverse RL and robotics datasets, converts control data of various data formats into a unified TensorFlow dataset format, and enables straightforward local storage and usage for training, fine-tuning, and evaluation.

\section{Evaluation Harness and Metric Suite}

To ensure the integrity of pre-training and fine-tuning processes, we introduce systematic test splits for the datasets within our collection as a part of our evaluation harness. These curated splits prevent data contamination and establish a reliable foundation for benchmarking, further enhancing the utility of MultiNet in advancing multimodal AI research.

We also introduce a carefully designed evaluation metrics suite to evaluate the generalization capabilities of SoTA VLMs and VLAs and capture their performance in a fair, quantifiable manner. This suite includes metrics such as Mean Squared Error, Brier Mean Absolute Error \cite{Brier1950}, Precision, Recall, F1 scores, Invalid output percentage to understand the performance of these models on expert offline trajectories of Reinforcement Learning and Robotics data. For multimodal understanding and generation capabilities, the evaluation suite includes metrics such as CIDEr \cite{vedantam2015ciderconsensusbasedimagedescription} to evaluate image captioning and image-based text retrieval, VQA accuracy for visual question answering, Recall@K for image understanding and text-based image retrieval, and Accuracy for commonsense reasoning and text understanding.

By making the test splits and implementation of evaluation metrics Open-Source we hope to enable the community to conduct evaluations of the generalization capabilities of VLMs and VLAs, understand the gaps in the path toward building generalist AI systems, and push the field forward.

\section{An Open-Source Universal Prompting Framework to Adapt VLMs}
\begin{figure}[H]
    \centering
    \includegraphics[width=1\linewidth]{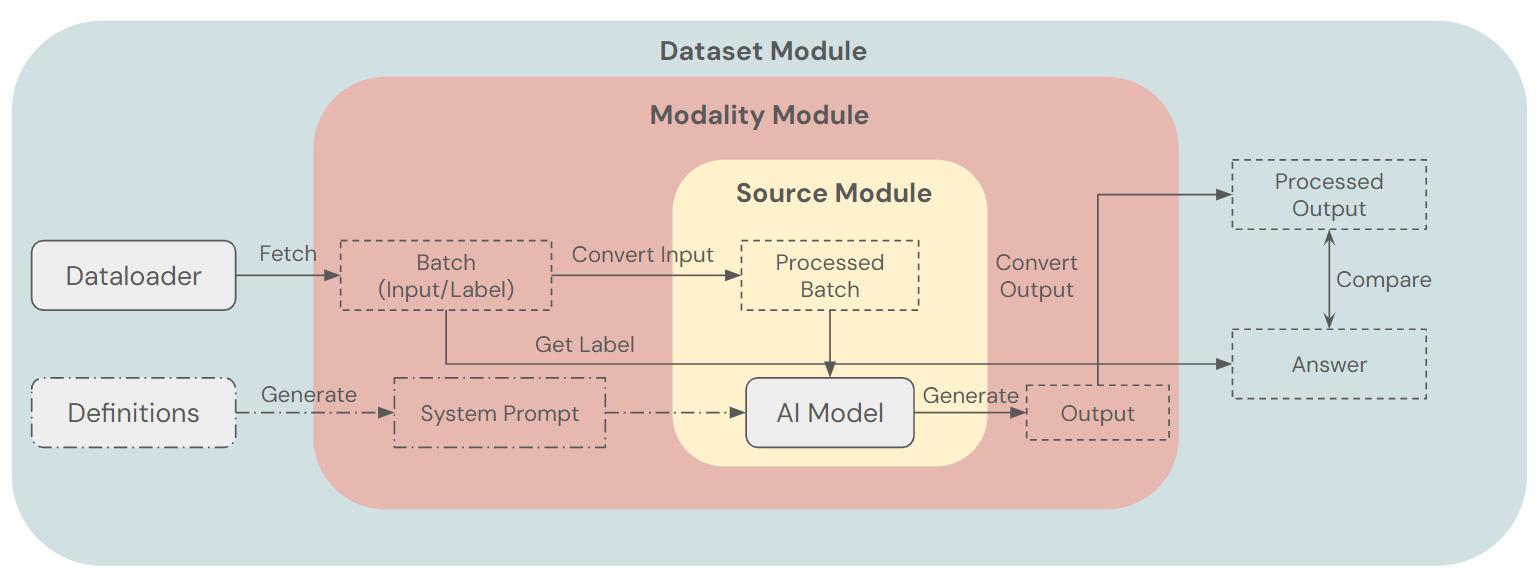}
    \caption{GenESIS is a modular framework that efficiently integrates diverse AI models and datasets into benchmark projects without disrupting existing components.}
    \label{fig:genesis_fig}
\end{figure}
We introduce GenESIS \footnote{https://github.com/ManifoldRG/MultiNet/tree/main/src/modules} seen in \ref{fig:genesis_fig}, a fully Open-Source, modular framework designed to streamline the integration of diverse VLMs across a multitude of tasks and datasets. GenESIS enables the efficient addition of new models and datasets to the MultiNet benchmark without disrupting existing components. It maximizes scalability and engineering efficiency through the following core principles:

\textbf{Interchangeability}: Models and datasets are designed to be mutually supportive and easily swappable, facilitating rapid experimentation and evaluation.

\textbf{Abstraction}: Common architectural patterns and logic are shared across modules, simplifying development, understanding, and testing. Base module classes significantly advance this principle by providing a standardized foundation for new components.

\textbf{Encapsulation}: Developers can focus on their specific model or dataset without requiring deep knowledge of unrelated modules, promoting modular development and reducing cognitive overhead.

\textbf{Prompt Engineering Framework}: A crucial intermediary that translates data of different domains (in this effort, robotics data) into a rich, structured textual representation, priming the VLM for analyzable outputs.

The GenESIS framework comprises the following key elements:

\textbf{System-Level Instructions}: Defines the overall goal, inference constraints, and the nature of the interaction (e.g., a simulated environment or robotics embodiment, etc.).

\textbf{Task and Environment Context}: Provides explicit descriptions of the specific task, its rules, and relevant environment details.

\textbf{Multimodal Input Integration}: Integrates visual observations within the prompt structure, enabling the VLM to contextualize its actions based on visual input.

\textbf{Action Space Definition}: Specifies the available actions, their format, and their corresponding verbal descriptions, ensuring clarity in the model's decision-making process.

\textbf{Output Instructions}: Defines the precise output structure, enabling the model to handle the complexity of generating outputs of very specific formats.

\section{Open-Source Architectural and Post Processing Adaptations for SoTA VLA Models}

Additionally, as part of MultiNet, we provide a set of SoTA VLA and VLM models that have been adapted to a variety of OOD domains, modalities, environments, and tasks. These adaptations involved processing model inputs and outputs, architectural changes, and inference pipelines to handle the specific structural and statistical characteristics of domains such as offline robotics embodiment datasets and procedurally generated discrete action environments. Our approach emphasizes modular, reusable, and Open-Source adaptations, enabling broader application and benchmarking of VLAs beyond their original training regimes.

\textbf{JAT} To adapt HuggingFace’s JAT model \cite{gallouédec2024jacktradesmastersome} to offline trajectories of robotics embodiment datasets, the inputs to the model were configured by pre-processing input image observations and concatenating floating-point observations into a singular tensor.

\textbf{OpenVLA} To adapt OpenVLA to offline trajectories of robotics embodiment datasets, the setup centered on standardizing gripper commands and managing action space compatibility. When adapted to procedurally generated, open-ended discrete action environments, the autoregressive step of the model was restricted to one for predicting single-dimensional action vectors. The generated actions were then unnormalized using statistical data derived from the complete Procgen subdataset. Finally, these unnormalized actions were rounded to obtain discrete action values. For evaluation, logits were extracted from the Llama 2 backbone \cite{touvron2023llama2openfoundation}, and probabilities were calculated. These probabilities were subsequently grouped by their associated integer action classes using cached mappings from Llama vocabulary tokens to Procgen integer actions.

\textbf{Pi0 Base}: To adapt the Pi0 Base model to procedurally generated discrete action environments, the action horizon was configured to a single timestep. The adaptation involved flow matching denoising over ten steps, utilizing a default action dimension of 32. The first dimension of the action output was selected, unnormalized using Procgen statistics, and then discretized by rounding to the nearest integer.

\textbf{Pi0 Fast}: To adapt the the Pi0 Fast model to procedurally generated discrete action environments the action horizon was set to one and autoregressive decoding was limited to four tokens. Probabilities were computed by mapping the backbone Paligemma \cite{beyer2024paligemmaversatile3bvlm} token IDs to Procgen actions. To mitigate its comparatively slower inference speed (approximately 10 times slower than OpenVLA), embeddings for static zero-images were cached after SigLIP \cite{zhai2023sigmoidlosslanguageimage} processing, which is the vision encoder model in the backbone. This optimization resulted in a doubling of the inference speed for Pi0 Fast.

\section{Research Results Using our Open-Source MultiNet Framework}

Our MultiNet framework was utilized to evaluate how SoTA VLMs, VLAs, and generalist models generalize to OOD data. The results show consistent failures to generalize to complex robotics and simulated digital action environments. Significant performance gaps highlight the importance of adaptable and platform-agnostic models. Factors such as training data distributions, architectural decisions, and processing techniques strongly influence the outcomes and lead to distinct behaviors in performance. We include some important results from these experiments in \ref{v01performance} and \ref{v02resultsmacrorecall}.

\section{Conclusion}

The MultiNet Benchmark, Framework, Toolkit, and Evaluation Harness are an early step towards the development of next-generation generalist AI systems. MultiNet and its associated tools are openly available, with the intention of empowering the broader research community, enhance reproducibility, and accelerate the collective pursuit of more capable and powerful generalist agents. We detail in future steps in  \ref{futurework}




\section*{Impact Statement}

This paper presents work whose goal is to advance the field of 
Machine Learning. There are many potential societal consequences 
of our work, none which we feel must be specifically highlighted here.


\bibliography{example_paper}
\bibliographystyle{icml2025}

\newpage
\appendix
\onecolumn
\section{Appendix}
\subsection{Future directions for MultiNet}
\label{futurework}
\begin{itemize}
    \item \textbf{Comprehensive Modality Analysis}: We will systematically investigate the interplay between control-task training and the emergent vision-language capabilities within Vision-Language Agents (VLAs). This exploration is critical for understanding architectural trade-offs and guiding the design of truly versatile, multi-modal generalist models.
    \item \textbf{Expanded Evaluation Horizons}: To rigorously assess and push the boundaries of generalization, MultiNet will integrate a more diverse array of control tasks, extending beyond current benchmarks like OpenX-Embodiment and Procgen. This expansion will allow for a thorough examination of model performance in entirely novel environments, pinpointing architectural bottlenecks and the frontiers of current generalization capabilities.
    \item \textbf{Advanced Transfer Learning Paradigms}: Our research will move beyond zero-shot evaluations to explore the efficacy of few-shot learning and fine-tuning strategies. A key focus will be on transferring learned skills to disparate domains, including complex software environments, thereby deepening our understanding of how to cultivate truly transferable representations across both embodied and digital tasks.
    \item \textbf{An Evolving Open-Source Simulation Benchmark}: We are actively working to transform MultiNet from its current offline evaluation framework into a dynamic, interactive, and open-source online benchmark. This will be powered by sophisticated world models underpinning state-of-the-art 2D and 3D simulation environments, enabling real-time assessment of agent capabilities in responsive and evolving settings. This open benchmark will serve as a community resource for robust and continuous evaluation.
    \item \textbf{Cultivating Multi-Domain Expertise}: A crucial research thrust involves developing advanced cross-domain adaptation mechanisms. The goal is to enable AI agents to seamlessly transfer and apply knowledge across a wide spectrum of environmental contexts, from robotic manipulation and navigation to software interaction and gameplay in diverse virtual worlds.
\end{itemize}
\newpage
\subsection{Dataset Spec}
\label{dataset_spec}


\begin{table}[h]
\centering
\caption{\small Overview of Datasets}
\label{tab:dataset_table}
\vskip 0.15in
\begin{center}
\begin{small}
\begin{sc}
\begin{tabular}{llc}
\toprule
\textbf{Dataset} & \textbf{Description / Task Type} & \textbf{Category} \\
\hline
OBELICS & Interleaved Image-Text & Vision-Language \\
COYO-700M & Image-Text pairs & Vision-Language \\
MS COCO & Object detection, segmentation, key-point detection, captioning & Vision-Language \\
Conceptual Captions & Image Captioning & Vision-Language \\
A-OKVQA & Visual Question Answering & Vision-Language \\
VQA-v2 & Visual Question Answering & Vision-Language \\
Datacomp-1B & Image-Text pairs & Vision-Language \\
Flickr30k & Image Captioning & Vision-Language \\
TextVQA & Visual Question Answering & Vision-Language \\
VizWiz & Visual Question Answering & Vision-Language \\
WinoGAViL & Vision-based Commonsense Reasoning & Vision-Language \\
ImageNet-R & Image-Text Pairs & Vision-Language \\
ObjectNet & Image-Text Pairs & Vision-Language \\
Fineweb-edu & High quality text corpus & Language \\
Hellaswag & Commonsense Reasoning & Language \\
ARC & Complex Reasoning and Knowledge Application & Language \\
CommonsenseQA & Commonsense Reasoning & Language \\
MMLU & Knowledge-intensive Question Answering & Language \\
DM Lab & Teach RL Agents 3D vision (Navigation-based control) & Control \\
DM Control Suite & Physics-based simulation environments (Locomotion-based control) & Control \\
ALE Atari & Atari games (Game-based control) & Control \\
Baby AI & Language-grounded navigation (Navigation-based control) & Control \\
MuJoCo & Multi-joint dynamics (Locomotion-based control) & Control \\
Meta-World & Meta-RL and Multi-task learning (Manipulation-based control) & Control \\
V-D4RL & Pixel-based analogues of DM Control Suite (Locomotion-based control) & Control \\
Procgen & Procedurally generated Atari-like environments (Game-based control) & Control \\
Open X Embodiment & Real-world Robotics tasks (Manipulation \& Locomotion control) & Control \\
LocoMuJoCo & Imitation learning for locomotion (Locomotion-based control) & Control \\
\bottomrule
\end{tabular}
\end{sc}
\end{small}
\end{center}
\vskip -0.1in
\end{table}

\subsection{MultiNet Benchmark Metrics}
\label{metrics}
\begin{table}[h]
\centering
\caption{\small MultiNet Benchmark metrics \& categories}
\label{tab:metric_table}
\vskip 0.15in
\begin{center}
\begin{small}
\begin{sc}
\begin{tabular}{llp{9cm}}
\toprule
\textbf{Metric} & \textbf{Evaluation Category} \\
\hline
Mean Squared Error & RL, Robotics \\
Brier Mean Absolute Error & RL, Robotics \\
Precision (Micro, Macro, and Class-wise variants) & RL, Robotics \\
Recall (Micro, Macro, and Class-wise variants) & RL, Robotics \\
F1 Score (Micro, Macro, and Class-wise variants) & RL, Robotics \\
Invalids Percentage & RL, Robotics \\
CIDEr & Image Captioning, Image-based Text retrieval \\
VQA Accuracy & Visual Question Answering \\
Recall@K & Image understanding, Text-based image retrieval \\
Accuracy & VQA, Commonsense reasoning, Text understanding \\
\bottomrule
\end{tabular}
\end{sc}
\end{small}
\end{center}
\vskip -0.1in
\end{table}


\newpage
\subsection{Model Performance on Out-of-Distribution Robotic Environments}
\label{v01performance}
\begin{figure}[h]
\centering
\includegraphics[width=\textwidth]{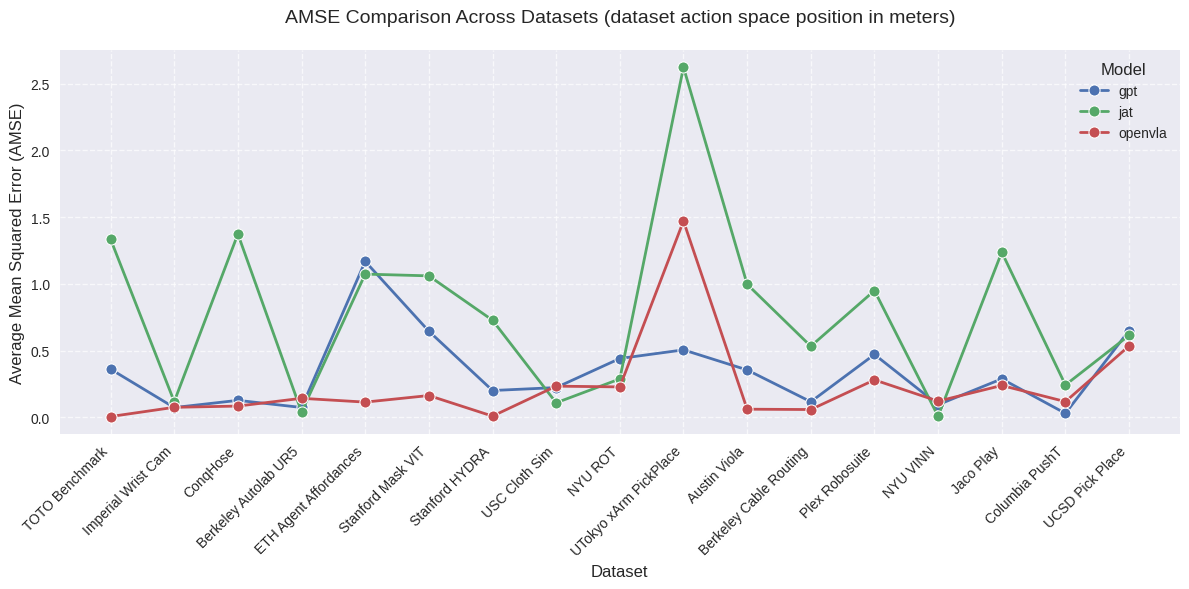}
\caption{AMSE values of GPT-4o, JAT, and OpenVLA Across 20 OpenX Datasets. JAT displays the poorest performance out of the 3 models with higher AMSE scores, while OpenVLA and GPT-4o demonstrate similar performance. OpenVLA displays consistent performance across most datasets.}
\label{v01results}
\end{figure}

\clearpage
\subsection{Model Performance on Out-of-Distribution Procedurally Generated Environments}
\label{v02resultsmacrorecall}
\begin{figure}[ht]
    \centering
    \includegraphics[width=10cm]{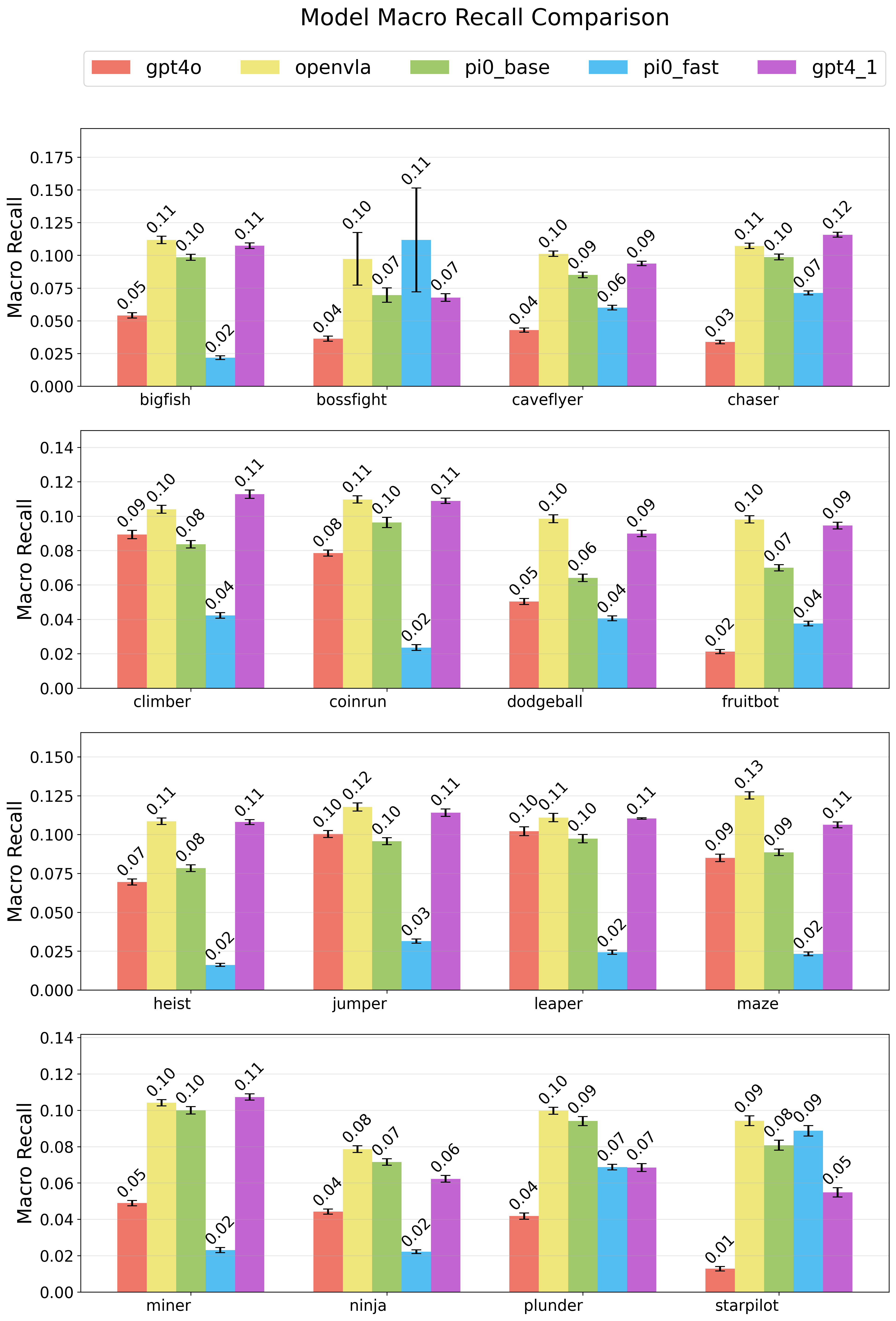}
    \caption{Macro recall across all 5 models. OpenVLA performs better when considering macro recall compared to macro precision, indicating a high number of false positives. GPT 4o shows lower macro recall than precision, indicating biased performance towards specific minority classes. GPT 4.1 and Pi0 Base show relatively less biased and moderate performance, whereas Pi0 FAST showed consistent low recall.}
    \label{fig:macro-recall}
\end{figure}
\begin{figure}
    \centering
    \includegraphics[width=10cm]{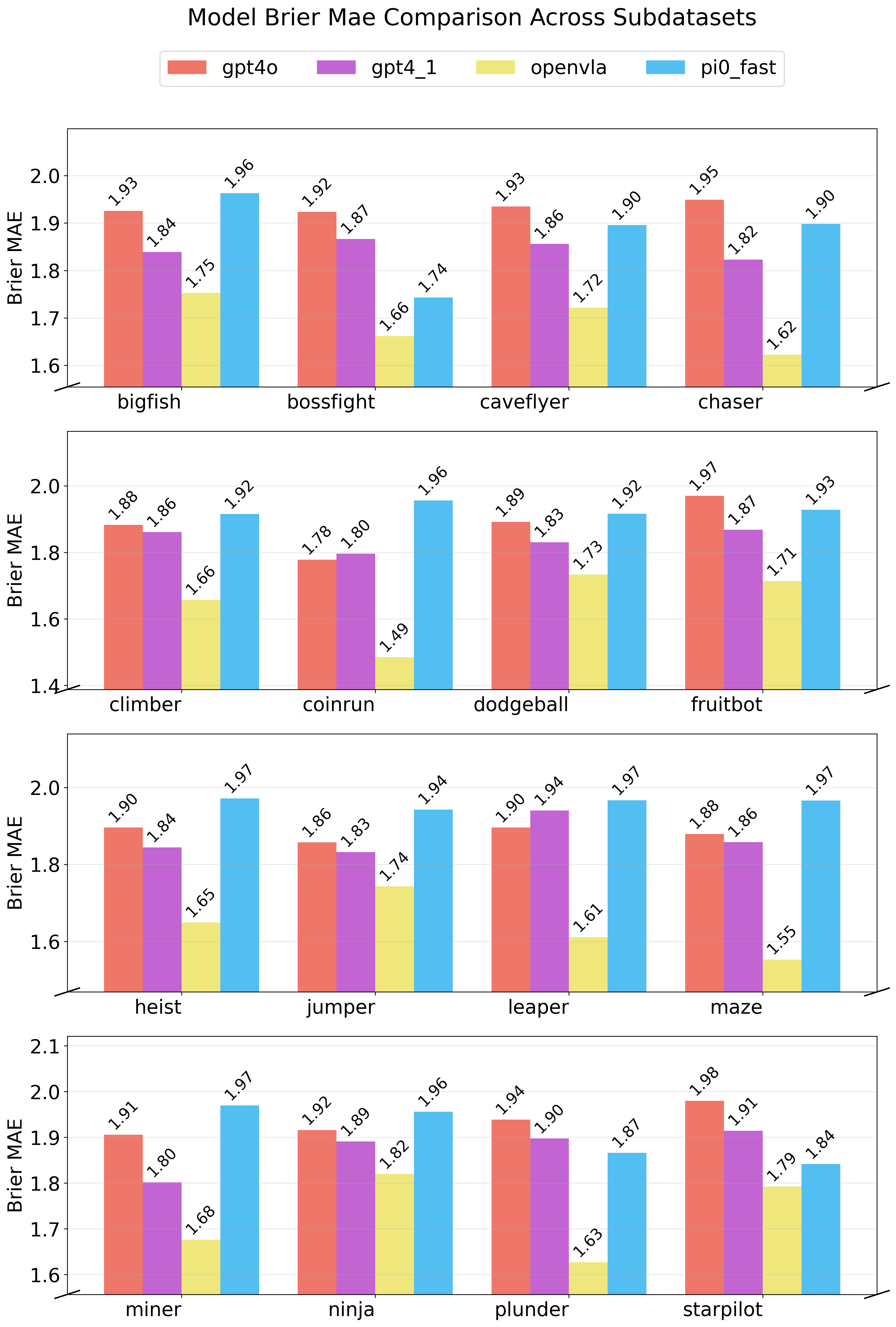}
    \caption{Brier Mean Absolute Error scores across 4 models - GPT 4o, OpenVLA, GPT 4.1, and Pi0 FAST. Pi0 Base is a diffusion-based model and can not be evaluated using Brier MAE due to a lack of logits in its inference architecture. All models display Brier MAE close to 2, indicating poor performance.}
    \label{fig:brier-mae}
\end{figure}
\begin{figure}
    \centering
    \includegraphics[width=10cm]{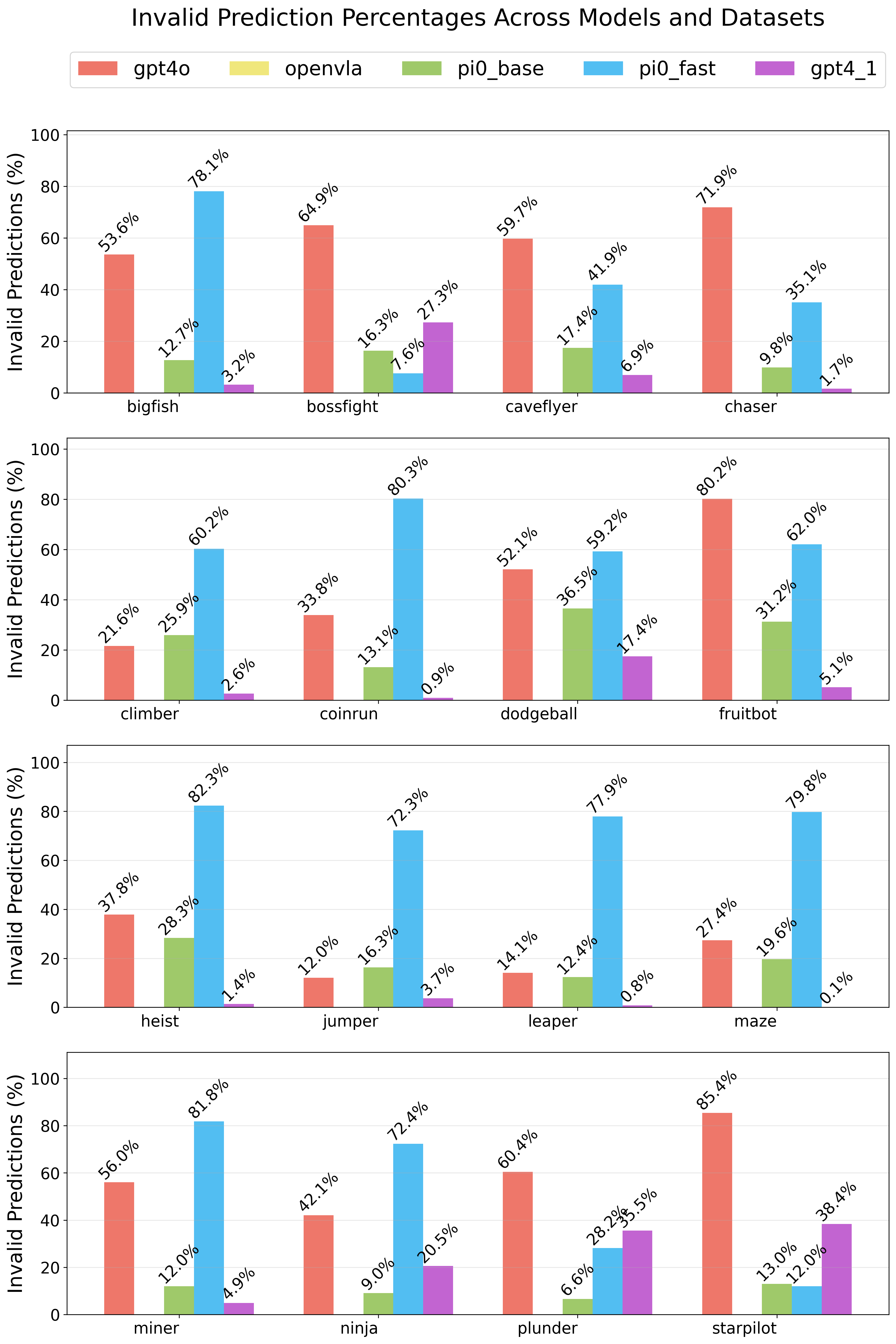}
    \caption{Percentage invalids across all 5 models. Invalids refer to model predictions that are not valid actions in the subdataset's action space. Pi0 FAST and GPT 4o struggle to produce valid actions irrespective of the subdataset.}
    \label{fig:invalid-percentage}
\end{figure}

\end{document}